\documentclass[%
 reprint,
 amsmath,amssymb,
 aps,
]{revtex4-2}

\usepackage{color}
\usepackage{comment}
\usepackage{graphicx}
\usepackage{dcolumn}
\usepackage{bm}
\usepackage{algorithm}
\usepackage{algorithmic}
\usepackage{bbold}

\begin{document}

\preprint{APS/123-QED}

\title{Learning physically consistent mathematical models from data using group sparsity}

\author{Suryanarayana Maddu$^{1,2,3,7}$, Bevan L.~Cheeseman$^{1,2,3,\dag}$, Christian L.~M\"{u}ller$^{4,5,6}$, Ivo F.~Sbalzarini$^{1,2,3,7,8}$} \email{sbalzarini@mpi-cbg.de}
\affiliation{ $^{1}$ Technische Universit\"{a}t Dresden, Faculty of Computer Science, 01069 Dresden, Germany}
\affiliation{$^{2}$ Max Planck Institute of Molecular Cell Biology and Genetics, 01307 Dresden, Germany}
\affiliation{ $^{3}$ Center for Systems Biology Dresden, 01307 Dresden, Germany} 
\affiliation{ $^{4}$ Center for Computational Mathematics, Flatiron Institute, New York, NY, USA}
\affiliation{ $^{5}$ Department of Statistics, LMU M\"{u}nchen, Munich, Germany}
\affiliation{ $^{6}$ Institute of Computational Biology, Helmholtz Zentrum M\"{u}nchen, Germany}
\affiliation{ $^{7}$ Center for Scalable Data Analytics and Artificial Intelligence ScaDS.AI, Dresden/Leipzig, Germany}
\affiliation{ $^{8}$ Cluster of Excellence Physics of Life, TU Dresden, Germany}
\altaffiliation{Now at: ONI Inc., Oxford, UK.}

\date{\today}

\begin{abstract}
We propose a statistical learning framework based on group-sparse regression that can be used to 1) enforce conservation laws, 2) ensure model equivalence, and 3) guarantee symmetries when learning or inferring differential-equation models from measurement data. Directly learning \textit{interpretable} mathematical models from data has emerged as a valuable modeling approach. However, in areas like biology, high noise levels, sensor-induced correlations, and strong inter-system variability can render data-driven models nonsensical or physically inconsistent without additional constraints on the model structure. Hence, it is important to leverage \emph{prior} knowledge from physical principles to learn ``biologically plausible and physically consistent'' models rather than models that simply fit the data best. We present a novel group Iterative Hard Thresholding (gIHT) algorithm and use stability selection to infer physically consistent models with minimal parameter tuning. We show several applications from systems biology that demonstrate the benefits of enforcing \emph{priors} in data-driven modeling.

\end{abstract}

\maketitle


\section{Introduction}

\noindent Mathematical modeling is fundamental to extracting governing principles of natural phenomena. Usually, mathematical models are formulated from first principles, such as symmetry relations and conservation laws. This classic approach of modeling natural systems has been successful in many domains of science amenable to mathematical treatment. However, in domains like biology, the success of first-principle modeling is limited~\cite{prost2015active, trepat2018mesoscale, popkin2016physics, sbalzarini2013modeling}. This is mostly attributed to the  ``complexity'' of biological systems where nonlinearity, stochasticity, multi-scale coupling, non-equilibrium behavior, self-organization, and emergent dynamics can dominate. Formulating mathematical models from first principles is difficult in complex or multi-scale systems, and the resulting models often have many unknown parameters.

Data-driven modeling has thus emerged as a complementary approach to first-principles modeling. Data-driven analysis and forecasting of complex systems was made possible by unprecedented advances in imaging and measurement technology, computing power, and algorithmic innovations. While purely data-driven models, like reservoir computing, can be very successful in predicting future behavior \cite{Pathak2018}, such ``black box'' models are often difficult to interpret for domain scientists. This raises the question how \emph{interpretable} mathematical models, such as ordinary or partial differential equations (ODE/PDE), can be learned directly from data. 

The idea of automatic inference of differential-equation models from data dates back to the field of time-series analysis~\cite{crutchfield1987equations, packard1980geometry}. Early works used least-squares fitting to estimates PDE coefficients from spatio-temporal data~\cite{vallette1997oscillations, bar1999fitting}. Many different approaches have since been proposed, e.g., Bayesian networks~\cite{daniels2015efficient}, linear dynamic models~\cite{friston2003dynamic}, recurrent neural networks~\cite{sussillo2009generating}, symbolic regression~\cite{schmidt2009distilling, schmidt2011automated},  sparse regression~\cite{brunton2016discovering, rudy2017data}, and artificial neural networks~\cite{raissi2017machine, raissi2019physics}. Methods based on sparse regression have been particularly successful, owing to their simplicity, computational efficiency, and applicability in the data-scarce regime \cite{maddu2019stability}. They have therefore found applications in many domains ranging from optics~\cite{sorokina2016sparse}, to plasma physics~\cite{dam2017sparse}, fluid mechanics~\cite{loiseau2017sparse}, chemical physics~\cite{hoffmann2019reactive}, aerospace engineering~\cite{el2018sparse}, and biology~\cite{maddu2019stability}. The sparse regression methodology has also been extended to incorporate control~\cite{brunton2016sparse}, implicit dynamics~\cite{mangan2016inferring}, parametric dependencies~\cite{rudy2019data}, stochastic dynamics~\cite{boninsegna2018sparse}, discrepancy models~\cite{de2019discovery}, and multi-scale physics~\cite{champion2019discovery}. Algorithms based on integral terms~\cite{schaeffer2017sparse}, automatic differentiation~\cite{both2019deepmod}, and weak formulations~\cite{reinbold2020using} have increased regression robustness by avoiding high-order derivatives of noisy data. All of these developments have corroborated the feasibility of data-driven learning of interpretable mathematical models. 

Given the feasibility of data-driven modeling, and the historic success of first-principles modeling, it seems natural to try combine the two. This requires methods to incorporate or enforce first-principle constraints, like conservation laws and symmetries, into the data-driven inference problem. First attempts in this direction used block-diagonal dictionaries with group sparsity to avoid model discrepancy~\cite{de2019discovery,Schaeffer2017b} and to infer PDEs with varying coefficients~\cite{rudy2019data}. However, there are many more \emph{priors} one may want to exploit when modeling complex systems, including information about possibilities of certain biochemical reactions, the presence of symmetries in interactions, knowledge of conservation laws, dimensional similarities, or awareness of spatially and temporally varying latent variables. Such prior knowledge can come from first principles or from model assumptions/hypotheses. To date, there is no statistical inference framework available that would allow flexible inclusion of different types of \emph{priors} into data-driven inference of differential equations models.

Here, we present a statistical learning framework based on group sparsity to enforce a wide range of physics and modeling \emph{priors} in the regression problem for robust inference of the structure of ordinary or partial differential equation (ODE/PDE) models. We present three representative examples from biology to demonstrate how information about conservation laws, latent variables, and symmetries can be encoded into grouped features of a sparse regression formulation. We therefore present numerical experiments using a mass-conserving ODE model for JAK-STAT signaling in cells, a mechanical transport model for membrane proteins, and $\lambda-\omega$ reaction diffusion systems, respectively. We solve the resulting non-convex optimization problems approximately using the group Iterative Hard Thresholding (gIHT) algorithm presented here, in combination with stability selection for statistically consistent model identification~\cite{maddu2019stability}. We show that stability selection in combination with gIHT enables robust model inference from limited, noisy data. 
 
\section{Problem Formulation}\label{section2}

\noindent We aim to learn the functional form of a governing ordinary or partial differential equation from data of the corresponding dynamics. We consider the following canonical form, where the right-hand side consists of a nonlinear function $\mathcal{N}$ of space $x$, time $t$, and derivatives:
\begin{equation}
    \frac{\partial u_i}{\partial t} = \mathcal{N} \left( x, t, \Xi(x,t), u_i, u_i u_j, \frac{\partial u_i}{\partial x_j}, \frac{\partial u_i u_j}{\partial x_j}, \frac{\partial^2 u_i}{\partial x_j^2 },... \right).
    \label{eq:1}
\end{equation}
The quantity $u_i$ is the state variable of interest (e.g., velocity, concentration, pressure) and $\Xi(x,t)$ is the set of parameters of the equation, like diffusion constants or viscosity. The dependence of $\Xi$ on $(x,t)$ allows for equations with varying coefficients  in both space and time. Common models like Navier-Stokes, advection, active mechano-chemistry, and reaction-diffusion models are represented by this canonical form. Models requiring a different left-hand side (e.g., wave equations) can be expressed using suitably adjusted canonical forms.

We follow the standard equation inference approach~\cite{rudy2017data, brunton2016discovering}, constructing an over-complete {\em dictionary} of possible right-hand side terms using discrete approximations (e.g., finite differences) of the derivatives from the data. For example, for a model with a single scalar state variable $u\in\mathbb{R}$, a dictionary of $p\in\mathbb{N}$ potential terms numerically evaluated over $N\in\mathbb{N}$ data points is a matrix $\bm{\Theta}\in \mathbb{R}^{N\times p}$. The canonical form of Eq.~\ref{eq:1} then becomes:
\begin{equation}\label{eq:explicit_govern}
\underbrace{\begin{bmatrix}
           \vert \\
           u_t \\
           \vert
\end{bmatrix}}_\text{$ \bm{U}_t(N\times 1)$} 
=
\underbrace{\begin{bmatrix}
          \:\: \vert \qquad \vert \qquad \vert \qquad \vert \qquad \vert \qquad \vert \quad\\
          u\quad uu_{x} \:  \:\:\: \ldots   u^3 u_{xx}  \: \ldots \:\:\: \ldots \\
          \:\: \vert \qquad \vert \qquad \vert \qquad \vert \qquad \vert \qquad \vert \quad
\end{bmatrix}}_\text{$\bm{\Theta} (N\times p)$}
 \underbrace{ \bm{\xi}}_{ (p \times 1)}.
\end{equation}
Here, we generally include in $\bm{\Theta}$ all differential operators and polynomial nonlinearities up to and including order 3. The left-hand side vector $\bm{U}_t(N\times1)$ is the discrete approximation to the temporal derivative at each data point, and each column of $\bm{\Theta}$ is the discrete approximation of one potential term of the right-hand side evaluated at all $N$ data points. $\bm{\xi}$ is the vector of unknown coefficients $\left[ \:\: \xi_0 \quad \xi_1 \quad \xi_2 \quad \xi_3 \quad \ldots \quad \xi_p \:\: \right]^{\top}$. 

The problem is to find a statistically consistent  $\bm{\xi}^{*}$ such that the model in Eq.~\ref{eq:explicit_govern} fits the data while being sparse, i.e., $\vert \bm{\xi}^{*} \vert_0 \ll p$. This trade-off between model complexity and data-fitting can be formulated as a regularized optimization problem:
\begin{equation}\label{general_opt}
    \hat{\bm{\xi}}^{\lambda} = \arg \min_{\bm{\xi}} \left( h\left( \bm{\xi} \right) + \lambda \: r \left( \bm{\xi} \right) \right),
\end{equation}
where $\hat{\bm{\xi}}^{\lambda}$ is the global minimizer, $h\left( \cdot \right)$ a smooth convex data-fitting function (e.g., least-squares or Huber loss), and $r\left( \cdot \right)$ a regularization or penalty function with tuning parameter $\lambda\in\mathbb{R}^+$ that controls the trade-off between model simplicity and fitting accuracy.

\section{Solution Method}

\noindent We provide an algorithm to solve the optimization problem in Eq.~\ref{general_opt} while accounting for modeling \emph{priors} and guaranteeing statistically stable and consistent models.

\subsection{Sparse regression}
\noindent To enforce sparsity, the problem in  Eq.~\ref{general_opt} is formulated as:
\begin{equation}\label{normalform}
\hat{\bm{\xi}}^{\lambda} = \arg\min_{\bm{\xi}} \frac{1}{2} \Vert \bm{U}_t -  \bm{\Theta} \bm{\xi} \Vert_{2}^{2} + \lambda  \Vert \bm{\xi} \Vert_0\, .
\end{equation}
The regularization $r\left( \bm{\xi} \right) = \lambda \Vert \bm{\xi} \Vert_0$ penalizes the number of non-zero terms in the model, hence favoring simpler models (Occam's razor) that are easier to interpret. Such sparsity-promoting regularization has very successful in applications of compressive sensing and signal processing. 

Algorithms that efficiently compute locally optimal solutions to Eq.~\ref{normalform} include greedy optimization strategies~\cite{tropp2004greed}, Compressed Sampling Matching Pursuit (CoSaMP)~\cite{needell2009cosamp}, subspace pursuit~\cite{dai2009subspace}, and Iterative Hard Thresholding (IHT)~\cite{blumensath2009iterative}. 

To avoid the problem of non-convexity in the objective function, a popular approach is to consider the convex relaxation of the problem in Eq.~\ref{normalform} by replacing the $\Vert \cdot \Vert_0$ term with $r(\bm{\xi}) = \Vert \bm{\xi} \Vert_1$ \cite{Tishbirani1996}. However, while this formulation benefits from the availability of fast convex optimization algorithms, it does not provide good approximations when model terms are correlated~\cite{yuan2006model} and leads to biased estimates of model coefficients~\cite{kowalski2014thresholding}, and thus yielding reduced model selection performance in practice \cite{maddu2019stability}. Therefore, we directly consider the original non-convex problem in Eq.~\ref{normalform} for model selection.

\subsection{Group sparse regression}\label{groupsparsereg}
\noindent We use the concept of group sparsity \cite{yuan2006model,Huang2010} to integrate modeling \emph{priors} into our sparse regression framework. 
We assume that prior knowledge about the underlying system can be expressed as a partitioning of model terms into $m$ groups. In the estimation process, group sparsity then encourages for the groups and their associated coefficients to enter or leave the statistical model \emph{jointly}.  
Formally, given a partitioning of the coefficients $\bm{\xi}_k, k = 1,2,\ldots ,p$, 
into $m$ groups $g_j, j = 1,2,\ldots ,m$, we consider the following objective:
\begin{align}\label{groupform}
\hat{\bm{\xi}}^{\lambda} = \arg\min_{\bm{\xi}} &\frac{1}{2} \Vert \bm{U}_t - \sum_{j=1}^{m}  \mathbf{\Theta}_{g_j} \bm{\xi}_{g_j} \Vert_{2}^{2} +\notag \\ 
&\lambda  \sum_{j=1}^{m} \sqrt{p_j} \: \mathbb{1}  \left ( \Vert \bm{\xi}_{g_j} \Vert_{2} \neq 0 \right),
\end{align}
where $\bm{\Theta}_{g_j} (N \times p_j)$ is the submatrix of $\bm{\Theta}$ formed by all columns corresponding to the coefficients in group $g_j \subseteq \{1,\ldots,p\}$ and $\bm{\xi}_{g_j} = \{ \bm{\xi}_i: i \in g_j \}$ is the coefficient vector $\bm{\xi}$ restricted to the index set $g_j$ of size $p_j$. Computing the indicator function $\mathbb{1}(\cdot)$ over the $\Vert \cdot \Vert_{2}$ norm encourages sparsity on the group level \cite{yuan2006model}. For groups comprising only a single element, the penalty reduces to the $\Vert\cdot\Vert_0$-norm.
Here, we restrict ourselves to \emph{non-overlapping} groups where $g_i \cap g_j = \emptyset$, $\forall i\neq j = 1,\ldots ,m$. Extensions to overlapping groups are possible \cite{jain2016structured} and discussed in section \ref{sec:concl}.

We solve the non-convex problem in Eq.~\ref{groupform} using a novel \emph{group Iterative Hard Thresholding} (gIHT) algorithm, which generalizes the standard IHT algorithm  and is detailed in the Appendix.



\subsection{Stability selection}
\noindent Robust tuning of the regularization parameter $\lambda$ is of fundamental importance for successful model discovery. Wrong choices of $\lambda$ result in incorrect equation models being identified, even if correct model discovery would, in principle, have been possible given the data. Common methods for tuning $\lambda$ include the Akaike information criterion (AIC)~\cite{akaike1998information}, the  (modified) Bayesian information criterion (BIC)~\cite{schwarz1978estimating}, and cross-validation. While AIC/BIC model selections is useful for combinatorial best-subset selection methods in low dimensions, they typically deteriorate in high dimensions since they rely on asymptotic considerations. Similarly, cross-validation tends to include many false-positive coefficients in the data-limited regime \cite{Lim2016}.

Here, we consider the statistical principle of stability selection, which tunes $\lambda$ so as to maximize model stability under sub-sampling of the data~\cite{meinshausen2010stability}. 
We perform stability selection by generating $B$ random
sub-samples $I_b^{*}$, $b=1,\ldots ,B$ of the data and using the gIHT algorithm to find the set $\hat{S}^{\lambda}[I_b^{*}] \subseteq \{1,\ldots,p\}$ of coefficients (or groups) for every data sub-sample $I^{*}$ for different $\lambda$ over the  {\em regularization path} $\Lambda=[\lambda_{\max}, \lambda_{\min}]$. Here, we choose $\lambda_\text{min} = 0.1 \lambda_\text{max}$. The probability that coefficient (or group) $j$ belongs to the selected subset for a given $\lambda$ is approximately
\begin{subequations}
\begin{align}
    \hat{\Pi}_{g_j}^{\lambda} &= \mathbb{P}[ g_j \cap  \hat{S}^{\lambda} \neq \emptyset ] \\
                            &\approx \frac{1}{B}\sum_{b=1}^{B} \mathbb{1} (g_j \cap \hat{S}^{\lambda}[I_{b}^{*}] \neq \emptyset), \quad g_j \subseteq \{1,...,p\}. \label{eq:importance_measure}
\end{align}
\end{subequations}
This is the {\em importance measure} for group or coefficient $j$. Plotting this as a function of $\lambda$ provides an interpretable way to assess the robustness of the estimation across levels of regularization in a so-called {\em stability plot}. 

To select a final model, stability selection chooses the set of stable coefficients (or groups) $\hat{S}_\text{stable} =\{j:\hat{\Pi}_{g_j}^{\lambda_{s}}>\pi_\text{th} \}$. The threshold $\pi_\text{th}$ is chosen to control the type I error of false positives~\cite{buhlmann2014high},
\begin{equation}\label{eq:bounds}
    \pi_\text{th} = \frac{1}{2} + \frac{\binom qk^2}{2\binom pk E_\text{fp}},
\end{equation}
where $E_\text{fp}$ is the upper bound on the expected number of  false positives, $q = \vert \hat{S}_\text{stable}\vert$, and $k$ is the group size. For a fixed value of $\pi_\text{th}$, we use this relation to find a $\lambda _s$ for which a given bound on the expected number of false positives, $E_\text{fp}$, is guaranteed. Throughout this work, we fix $\pi_\text{th} = 0.8$ and $E_\text{fp} = 1$. Alternatively, one can determine $\pi_\text{th}$ by visual inspection of a stability plot, which usually shows clear separation between two groups of coefficients of different stability.

Stability selection not only removes the necessity to manually tune $\lambda$, but it also ensures robustness against data sampling and noise in the data. All of these properties are required for statistical consistency in the sense that the inferred models are guaranteed to become accurate with high probability with increasing data size~\cite{daniels2015efficient}.

\section{Applications}
\noindent We present three different modeling examples from systems biology that illustrate the utility of  \emph{priors} in data-driven modeling. Each example highlights a different type of \emph{prior} knowledge to be enforced. To emulate noisy measurements from real-world experiments, we corrupt the simulation data $u \in \mathbb{R}^N$ with additive Gaussian noise as follows: $\hat{u} = u + \sigma \cdot \mathcal{N}(0, \textrm{std}(u))$, where $\sigma$ is the level of the Gaussian noise added. We use polynomial differentiation to compute the spatial and temporal derivatives used to construct the dictionary.

\begin{figure}[!t]
\centering
\includegraphics[width=2.5in]{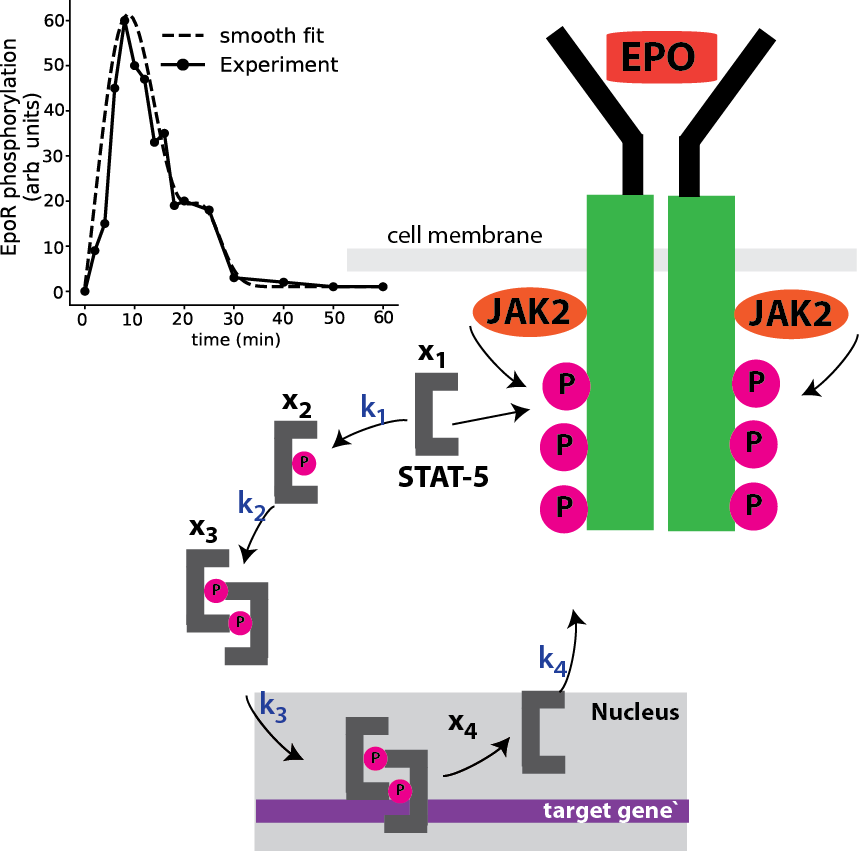}
\caption{\footnotesize{ \textbf{The core module of the JAK-STAT signaling pathway}. The hormone EPO binding to the EpoR receptor results in activation of the receptor (activated form EpoRa with concentration $c(t)$) by transphosphorylation of JAK2 and subsequently in tyrosine phosphorylation (P) of JAK2 and the EpoR cytoplasmic domain. Phosphotyrosine residues 343 and 401 in EpoR mediate recruitment of monomeric STAT-5 (concentration ${x}_1$). Upon receptor recruitment, monomeric STAT-5 is tyrosine phosphorylated (${x}_2$), dimerizes (${x}_3$), and migrates to the nucleus (${x}_4$), where it binds to the promoter of target genes and is dephosphorylated and exported again to the cytoplasm~\cite{timmer2004modeling}. The inset plot shows a time courses of EpoR activation as measured experimentally (data from \cite{swameye2003identification}).}
}
\label{fig:JAK_STAT}
\end{figure}

\subsection{Enforcing mass conservation in the JAK-STAT reaction pathway for signal transduction}
\noindent Signal transduction pathways are the engines of chemical information processing in living biological cells. Using methods from biochemistry and systems biology, the constituent molecules of many signalling pathways have been identified. Yet, identifying the topology of these chemical reaction networks remains challenging. 
It typically involves building mathematical models of hypothetical reaction networks and comparing their predictions with the data. A popular choice is to use ordinary differential equation (ODE)  models of the stoichiometry and chemical kinetics of the pathway. However, when discrepancies occur between the ODE model and the experimental data, it is difficult to decide whether the model structure is incorrect or whether the parameters of the model have been badly chosen~\cite{timmer2004modeling}. Here, data-driven modeling can help identify the stable structure of minimal ODE models that can explain the measurement data.

In this example, we consider the JAK-STAT pathway, which communicates chemical signals from outside a biological cell to the cell nucleus. It is implicated in a variety of biological processes from immunity to cell division, cell death, and tumour formation. Mathematical models based on biochemical knowledge of the JAK-STAT pathway have identified nucleo-cytoplasmic cycling as an essential component of the JAK-STAT mechanism, which has been experimentally verified~\cite{timmer2004modeling, swameye2003identification}.
We therefore consider the simplest ODE model with irreversible reactions that account for nucleo-cytoplasmic cycling and model information transfer from the cell membrane to the nucleus as previously discussed~\cite{timmer2004modeling}:
\begin{subequations}
\label{JKSTAT}
\begin{align}
    \dot{x}_1(t) &= -k^-_1  x_1(t) c(t) + 2 k^+_4 x_4(t), \label{eq:x1}\\
    \dot{x}_2(t) &= +k^+_1  x_1(t) c(t) - k^-_2 x_2^2(t), \label{eq:x2}\\
    \dot{x}_3(t) &= -k^-_3  x_3(t) + \frac{1}{2} k^+_2 x_2^2(t), \label{eq:x3}\\
    \dot{x}_4(t) &= +k^+_3  x_3(t) - k^-_4 x_4(t). \label{eq:x4}
\end{align}
\end{subequations}
The schematic for the JAK-STAT is shown in Fig.~\ref{fig:JAK_STAT} which illustrates the reaction cascade from outside the cell membrane to inside the cell nucleus. The functions $x_1(t)$, $x_2(t)$, $x_3(t)$, and $x_4(t)$ are the time courses of the concentrations of monomeric STAT-5, phosphorylated STAT-5, cytoplasmic dimeric STAT-5, and STAT-5 in the nucleus, respectively. The scalar constants $k_1^{\pm}$, $k_2^{\pm}$, $k_3^{\pm}$, and $k_4^{\pm}$ are the kinetic reaction rates of phosphorylation, dimerization, nuclear transport, and nuclear export, respectively. While of course $k^-_1=k^+_1$, $k^-_2=k^+_2$, $k^-_3=k^+_3$, and $k^-_4=k^+_4$, we distinguish different occurrences of the same rate constant by sign superscripts in order to make clear that they are learned from data independently by our regression algorithm. 

\begin{figure}[!t]
\centering
\includegraphics[width=3.2in]{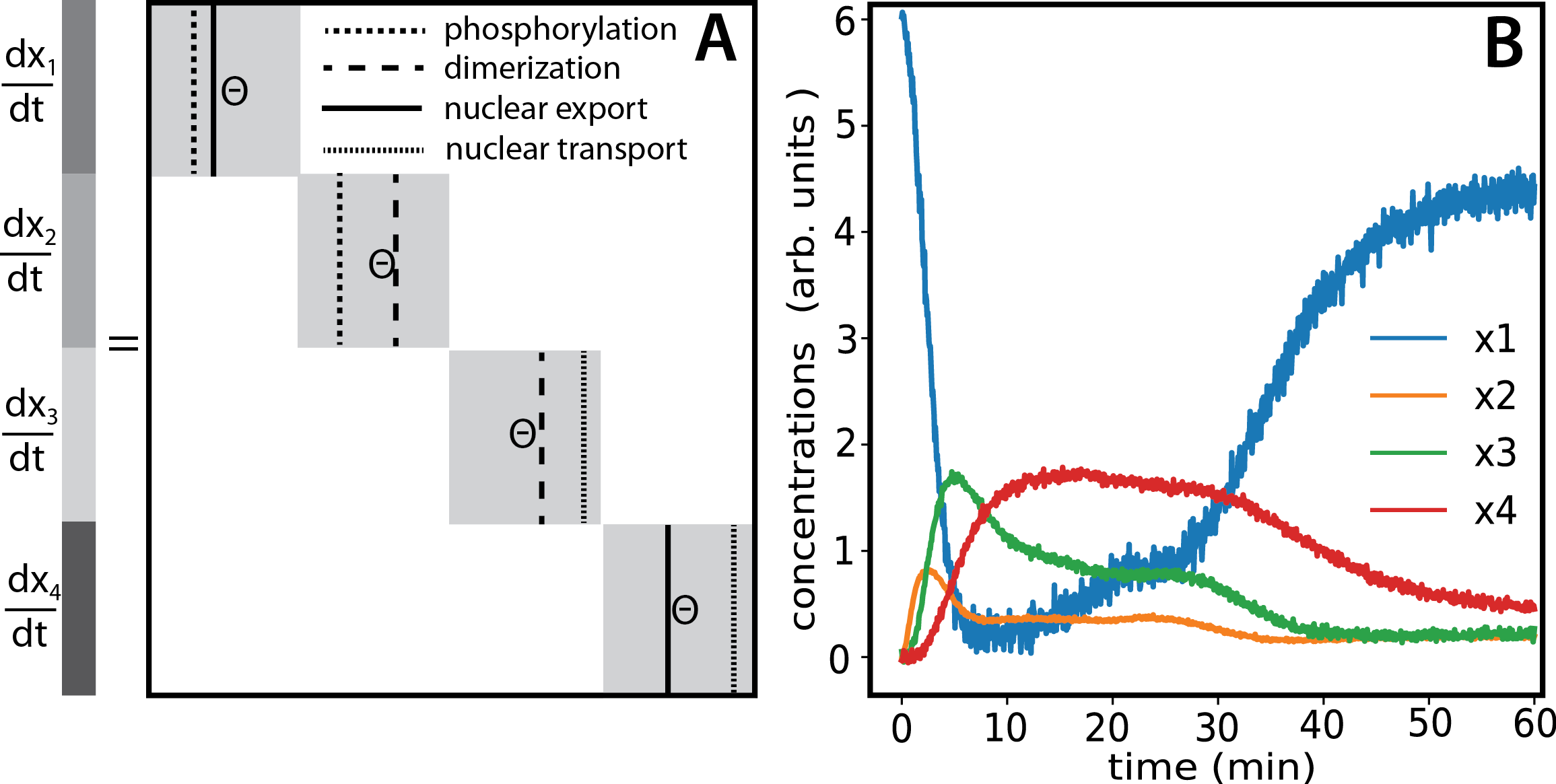} 
\caption{\textbf{Dictionary design and simulated data.} \footnotesize{ \textbf{A}) Dictionary construction and coefficient grouping. The identical dictionary $\bm{\Theta}$ for each component is stacked in a block-diagonal matrix for joint learning. Vertical lines indicate the four coefficient groups $g_1, \ldots , g_4$, corresponding to the four biochemical processes named in the legend. \textbf{B}) The time-series data for different concentrations in the JAK-STAT pathway obtained by numerically integrating the deterministic ODE model in Eqs.~(\ref{JKSTAT}) using the ode45 \textrm{MatLab} solver.
The time-series data is then corrupted with 10\% additive Gaussian noise ($\sigma=0.1$).}}
\label{JAK_STAT_dict}
\end{figure}

For sparse-regression model learning, a dictionary matrix $\bm{\Theta}$ of all possible interactions between the molecules is generated (see Eq.~\ref{eq:explicit_govern}). The left-hand side $\bm{U}_t$ is the time derivative of each concentration, i.e., $\dot{{x}}_1$, $\dot{{x}}_2$, $\dot{{x}}_3$, and $\dot{{x}}_4$ as approximated from the data. For this application, $\bm{\Theta}$ contains $p=19$ polynomial nonlinearities (e.g., ${x}_1, {x}_2, {x}_1^2, {x}_1 {x}_2, {x}_1 {x}_2 {x}_3, \ldots$), corresponding to chemical kinetics of different orders. The same $\bm{\Theta}_i = \bm{\Theta}$ is used for each component ${x}_i$, $i=1,2,3,4$, leading to the block-diagonal overall structure shown in Fig.~\ref{JAK_STAT_dict}A. For model inference, we use the simulated concentration time-courses shown in Fig.~\ref{JAK_STAT_dict}B. They are obtained from numerically solving the model Eqs.~(\ref{JKSTAT}) with $k^-_1=k^+_1 = 0.021$, $k^-_2=k^+_2 = 2.46$, $k^-_3=k^+_3 = 0.2066$, and $k^-_4=k^+_4 = 0.10658$ as found by fitting experimental data~\cite{timmer2004modeling, swameye2003identification}, see Fig.~\ref{fig:JAK_STAT} inset. The simulated data are corrupted by 10\% additive Gaussian noise before inference. 
The noisy time-series data for the activated Epo receptor, $c(t)$, is taken directly from  experimental measurements \cite{swameye2003identification}. All units are relative to the experimental data.

Using the simulated data, we aim to infer back the model equations. The JAK-STAT pathway conserves mass, as evident from the ODE model Eqs.~(\ref{JKSTAT}). 
This can be used as a \emph{prior} when inferring a model from data. We therefore perform group-sparse regression (see Sec.~\ref{groupsparsereg}) using the groups
\begin{align}
g_1 &= \{i : \textrm{column index of }  {x}_1 \textrm{ in }  \mathbf{\Theta}_1, \mathbf{\Theta}_2  \}, \\
g_2 &= \{i : \textrm{column index of }  {x}_{2}^{2} \textrm{ in }  \mathbf{\Theta}_2, \mathbf{\Theta}_3  \}, \\
g_3 &= \{i : \textrm{column index of }  {x}_3 \textrm{ in }  \mathbf{\Theta}_3, \mathbf{\Theta}_4  \}, \\
g_4 &= \{i : \textrm{column index of }  {x}_4 \textrm{ in }  \mathbf{\Theta}_1, \mathbf{\Theta}_4  \}.
\end{align}

\begin{figure}[!t]
\centering
\includegraphics[width=3.2in]{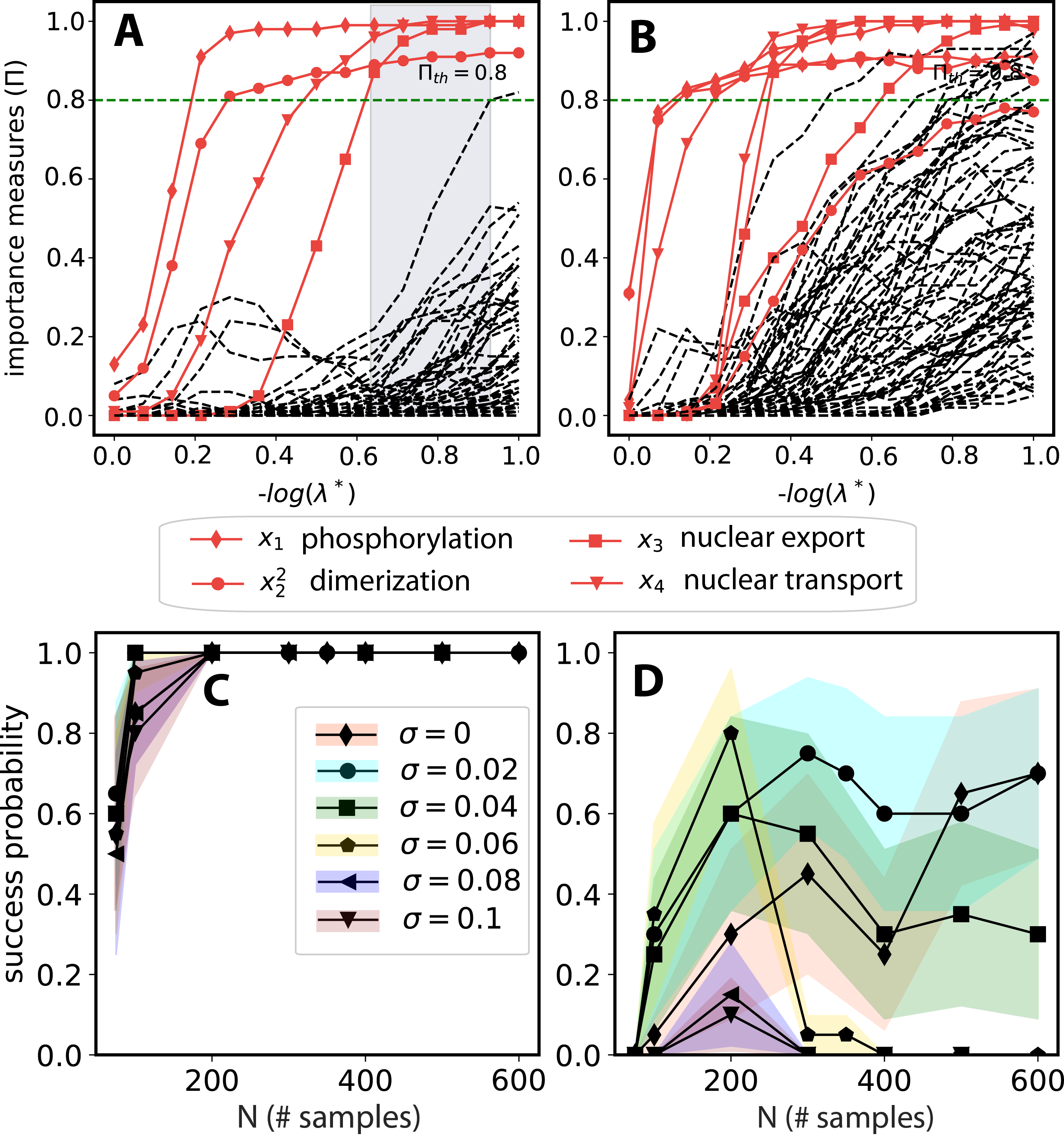} 
\caption{\textbf{Inferring JAK-STAT signalling models from noisy data.}\footnotesize{ \textbf{A}) Stability selection using grouping based on mass conservation. In the gray shaded range of $\lambda$ values, stability selection with $\Pi_{k}^{\lambda} \geq 0.8$ identifies the correct model. The solid red lines show the behavior of the true components of the ODE model, the dashed black lines are all other dictionary terms ($p=19$). \textbf{B}) Stability selection without grouping. There is no value of $\lambda$ for which the true model is found. In both (A) and (B), additive noise with $\sigma=0.1$ was added to the simulation data before inference and $N=200$ time points are used. \textbf{C}) Achievability plot for model selection with mass conservation \emph{prior}. \textbf{D}) Achievability plot for model selection without mass conservation \emph{prior}. In (C,D) the success probability of inferring the correct model over 20 independent trials are shown as a function of the number of data points used. Colored bands are Bernoulli standard deviations for different amounts of noise added to the simulation data prior to inference (see inset legend).}}
\label{JAK_STAT_SP}
\vspace{-1em}
\end{figure}

\noindent This is graphically represented by the vertical lines in Fig.~\ref{JAK_STAT_dict}A, with each group corresponding to one type of biochemical process in the model, as given in the inset legend. We solve the resulting group-sparse regression problem using our gIHT algorithm. This leads to a conservative model {\em structure}, but the fitted {\em values} of the rate constants may differ for different signs, i.e., it can be $k^-_1 \neq k^+_1$, etc. Enforcing symmetry also in the coefficient values, and not only in the model structure, would require solving a {\em constrained} group-sparse regression problem, which we do not consider here. 

The results are shown in Fig.~\ref{JAK_STAT_SP}A. In this benchmark setting, group sparsity helps identify the correct model terms (red curves) out of all terms of the dictionary. There exists a range of $\lambda_s$ values where stability selection with threshold $\pi_\text{th}=0.8$ (green dashed line) can identify the correct model, even at the 10\% noise level considered here.

Without coefficient grouping, i.e.~without mass-conservation \emph{prior}, there is no value of $\lambda$ for which the correct model is recovered, as shown in Fig.~\ref{JAK_STAT_SP}B. To show consistency of the group-sparsity method, we also provide achievability plots in Fig.~\ref{JAK_STAT_SP}C,D. They show that enforcing the mass conservation \emph{prior} leads to consistent model selection over a wide range of sample sizes ($N$).

Using group sparsity in combination with stability selection, the correct model can be identified in 100\% of cases (over 20 independent repetitions) when $N>200$ data points are used (i.e., success probability 1), regardless of the noise level in the data (color, see inset legend), as shown in Fig.~\ref{JAK_STAT_SP}C. Sparse regression without \emph{priors} suffers from inconsistency, at all noise levels and for all data sizes (Fig.~\ref{JAK_STAT_SP}D). 
The learned coefficients at different noise levels are shown in Fig.~\ref{fig:JK-STATcoeff} in the Appendix.


\subsection{Enforcing model equivalence in advection-diffusion models of protein transport}
\begin{figure}[!t]
\centering
\includegraphics[width=3.4in]{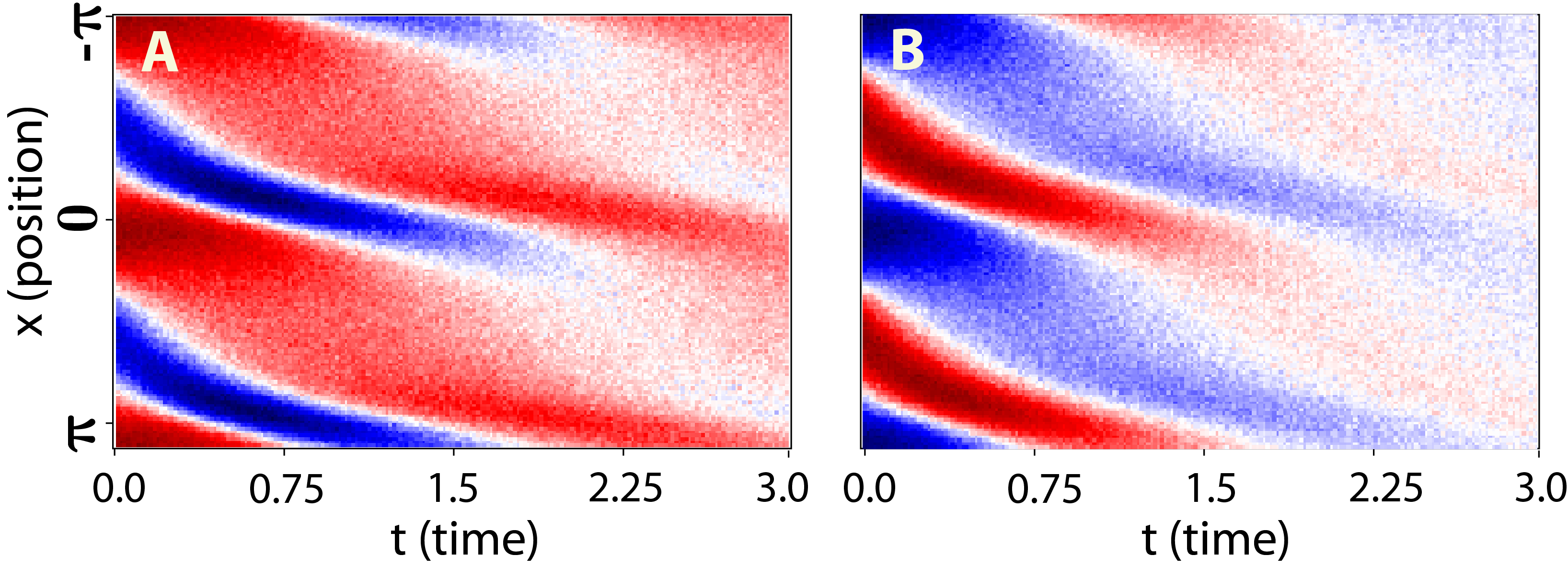}
\caption{\textbf{Simulation data used to learn spatio-temporal models of 1D advection-diffusion dynamics. A,B})
\footnotesize{Visualization of the data for $u(x,t)$ (in A) and $v(x,t)$ (in B) with 15\% additive Gaussian noise ($\sigma=0.15$). Spatial and temporal discretization uses 256 and 200 regularly spaced grid points, respectively. The solution is obtained via spectral differentiation and fourth-order Runge-Kutta time integration.
The diffusion constants of the species are $D_u = 0.25$ and $D_v = 0.5$ in non-dimensional units. The equations are solved with periodic boundary conditions for the time horizon $t \in [0, 3]$ with initial conditions $u(x,t=0) = \cos(\frac{2\pi x}{L})$, $v(x,t=0) = -\cos(\frac{2\pi x}{L})$ for species $u$ and $v$, respectively. The spatially varying velocity $c(x) = -\frac{3}{2} + \cos(\frac{2\pi x}{L})$ is used to advect the species.}} 
\label{AD_data}
\end{figure}

\noindent 
The development of organisms from their zygotic state involves a myriad of biochemical interactions coupled with the mechanical forces that eventually shape the resulting tissue. 
In the past decades, the role of mechanics, including forces and flows, has increasingly been investigated in developmental biology and morphogenesis. On the cell and tissue scale, many developmental processes involve both patterning and flows, including polarity establishment, tissue folding, and cell sorting \cite{mayer2010anisotropies, mammoto2010mechanical}. Fluorescence imaging techniques enable quantification of the spatio-temporal concentration fields of labeled proteins~\cite{goehring2011polarization,gross2019guiding}. This has led to quantitative measurements and predictive models of active mechano-chemical self-organization of, e.g., as cytoplasmic flow \cite{nazockdast2017cytoplasmic}, endocytosis \cite{collinet2010systems}, and tissue patterning \cite{eaton2011cell}.

In this example, we consider simplest case of transport by advection and diffusion of signaling molecules. In order to allow for latent processes, we consider spatially varying model coefficients. We construct groups that allow the advection velocity (coefficients) to be a function of space. In addition, we also impose a \emph{prior} that promotes model equivalence, i.e., learning structurally similar models for the different chemical species, albeit with different diffusion constants. For the concentration fields $u(x,t)$ and $v(x,t)$ of two chemicals, this amounts to the model
\begin{align}
    \frac{\partial u}{\partial t} + c(x) \frac{\partial  u }{\partial x} + u \frac{\partial c(x)}{\partial x} &= D_u \frac{\partial^2 u}{\partial x^2}, \\ 
    \frac{\partial v}{\partial t} + c(x) \frac{\partial  v }{\partial x} + v \frac{\partial c(x)}{\partial x} &= D_v \frac{\partial^2 v}{\partial x^2} .
\end{align}
Here, $D_u, D_v$ are the respective diffusion constants, and the function $c(x)$ is the spatially varying advection velocity field. With added chemical reactions, this form of model has previously been successfully used to explain early patterning in the single-cell \emph{C.~elegans} zygote~\cite{goehring2011polarization, gross2019guiding}.

\begin{figure}[!t]
\centering
\includegraphics[width=3.4in]{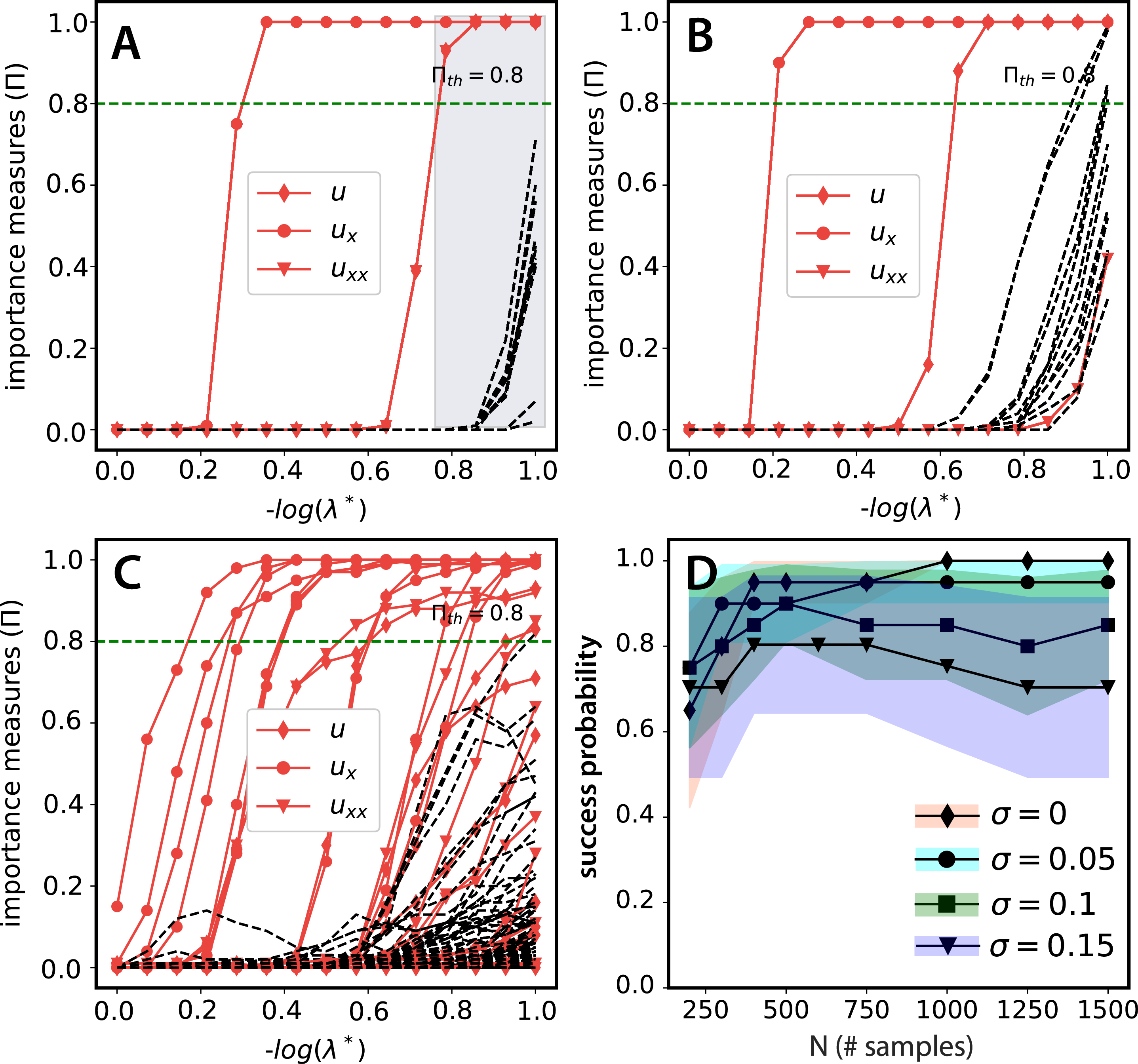}
\caption{\textbf{Inferring advection-diffusion dynamics with unknown spatially varying velocity field. }\footnotesize{\textbf{A}) Stability selection with groups to encode both spatially varying coefficients and model equivalence between the species. The gray shaded region is the range of $\lambda$ for which model selection with $\pi_{th} = 0.8$ identifies the correct model. \textbf{B}) Stability selection with groups only to encode spatially varying coefficients, but no grouping for model equivalence. \textbf{C}) Stability selection with no groupings at all. The solid red lines correspond to the true components of the PDE, with symbols referring to the differential operators as given in the inset legends. In A, B, C the dictionary size is $p=15$ and $p_g=10$ with 15\% Gaussian noise ($\sigma$
=0.15) added to the simulation data. \textbf{D}) Achievability plot for model selection using both \emph{priors} with $p_g=10$ and different levels of noise in the data.
Each point is averaged over 20 independent trials. The colored bands correspond to the Bernoulli variance.
}}
\label{fig:PAR_stability}
\end{figure}

We use data from numerical simulations of the above model equations with 15\% additive Gaussian noise (see Fig.~\ref{AD_data}) to show that both \emph{priors}, model equivalence and spatial variability, are necessary to recover the ground-truth equations including the spatially varying velocity field. We again first construct two block-diagonal dictionaries, for $u$ and $v$, where each block represents the dictionary constructed at one spatial location. 
We use $p_g = 10$ blocks (number of spatial points sampled), corresponding to 10 randomly selected spatial data points. Each of the diagonal blocks $\Theta (N,p)$ uses $N = 75$ randomly chosen time points and $p = 15$ potential operators.

We use grouping to enforce that the structure of the model learned from the data must be the same for all spatial locations, and that the models learned for $u$ and $v$ must be equivalent. Each group therefore ties a column in a block dictionary to all corresponding columns in the other blocks. This construction results in the following groupings to encode spatial variability:
\begin{equation}\label{eq:grouping_1}
    g_{l} = \{\{l + k \: p \}: \forall \:\: k \in \{ 0,..., p_g-1\} \}.
\end{equation}
\noindent Here, the set $g_l$ is the group $l$ and $p$ is the number of columns of the block dictionary. The group sets  $g_l^u$ and $g_l^v$, constructed for species $u$ and $v$ using Eq.~(\ref{eq:grouping_1}), can further be combined to enforce model equivalence between species with the grouping: $g_{l} = g_{l}^{u} \cup g_l^{v}$.

The resulting stability and achievability plots are shown in Fig.~\ref{fig:PAR_stability} when using the noisy data from Fig.~\ref{AD_data}. Comparing Fig.~\ref{fig:PAR_stability}A and B, we see that the \emph{prior} for model equivalence is necessary to recover the true model. The algorithm is unable to identify the diffusion process of the species $u$ when only using the grouping for the spatially varying coefficient  (Fig~\ref{fig:PAR_stability}B). Inference without any \emph{priors} fails to recover the true model even for noise-free data (Fig~\ref{fig:PAR_stability}C). The achievability plot in Fig.~\ref{fig:PAR_stability}D demonstrates the consistency of our model selection algorithm with grouping over 20 independent realizations of the noise process and random sub-sampling of the data. We observe consistent model recovery with high success probability even at high noise levels, albeit with decreasing fidelity as seen in Fig~\ref{fig:PAR_stability}D. In contrast, previous studies on advection-diffusion model recovery with unknown velocity field were limited to 1\% noise ($\sigma=0.01$)~\cite{rudy2019data}. 

The estimated latent velocity fields and their gradients are shown in Appendix Fig.~\ref{fig:PARprofiles} and compared with ground truth for different noise levels. 

\subsection{Enforcing symmetry in reaction-diffusion kinetics}
\begin{figure}[!t]
\centering
\includegraphics[width=3.4in]{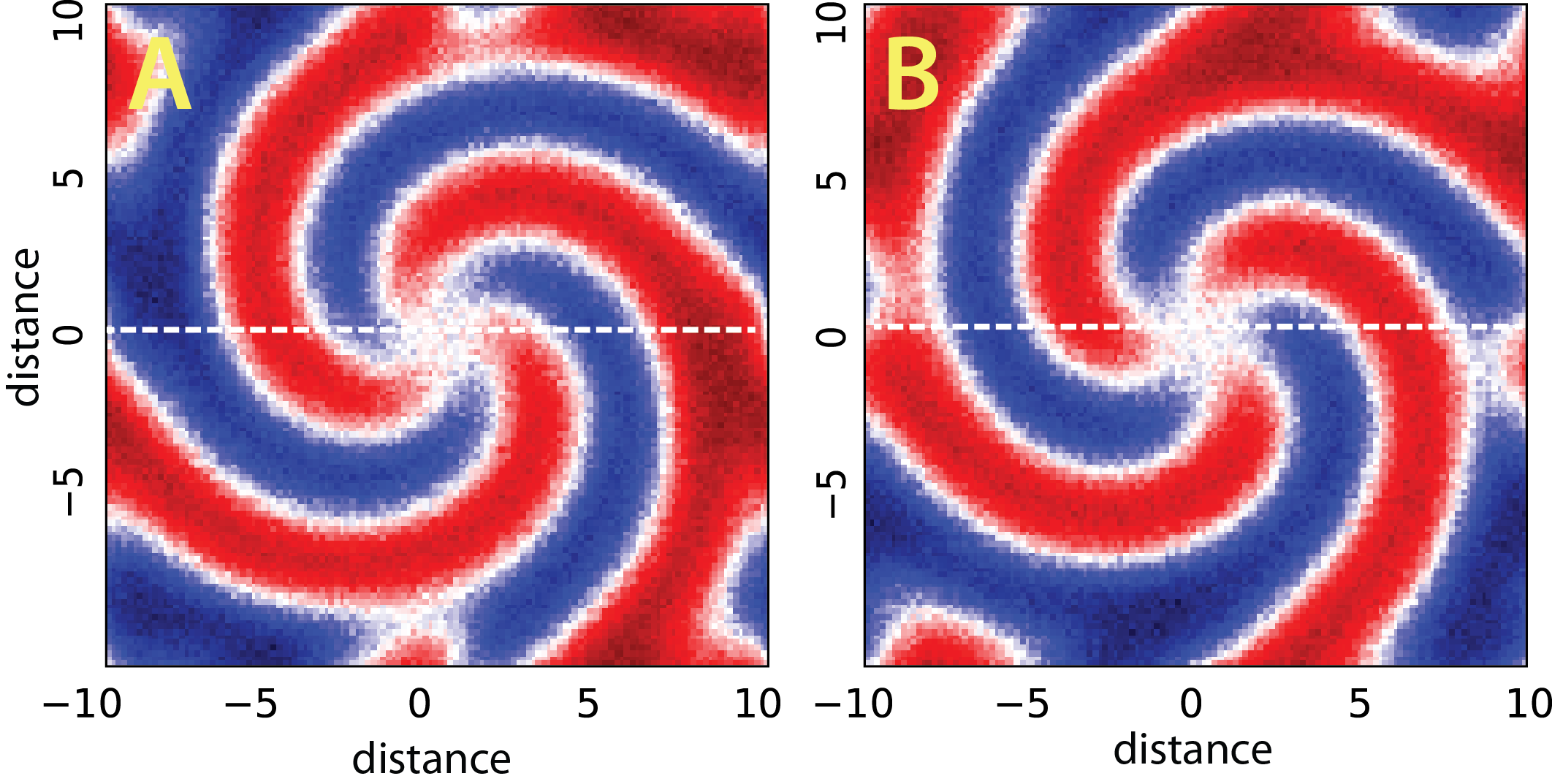}
\caption{\textbf{Simulation data used to learn reaction-diffusion dynamics.}\footnotesize{ \textbf{A,B})
Visualization of the 2D concentration fields $u(x,y)$ (in A) and $v(x,y)$ (in B) at time $t=7.5$ from numerical solution of the model with 10\% additive Gaussian noise ($\sigma=0.1$). The solution is obtained via spectral differentiation and fourth-order Runge-Kutta time integration with step size $dt=0.05$ on a Cartesian grid of $128 \times 128$ points with initial conditions $u(x,y,0) = \textrm{tanh} \left( \sqrt{x^2 + y^2} \cos \left( 3 \angle(x + iy) -  \sqrt{x^2 + y^2} \right)\right)$, and $v(x,y,0) = \textrm{tanh} \left( \sqrt{x^2 + y^2} \sin \left( 3 \angle(x + iy) -  \sqrt{x^2 + y^2} \right)\right)$.
\vspace{-2.em}
}}
\label{rd-data}
\end{figure}

\noindent Reaction-diffusion models are widely used in systems biology to describe the dynamics of chemical reaction networks in a homogeneous space. Their popularity goes back to a seminal paper by Alan Turing \cite{turing1952}, proposing that reaction-diffusion mechanisms could be responsible for pattern formation
in developing tissues. Since then, reaction-diffusion equations have been successful in modeling non-equilibrium pattern formation~\cite{cross1993pattern}, dynamics of ecological and biological systems~\cite{murray2007mathematical, medvinsky2002spatiotemporal}, cell polarity~\cite{cross1993pattern, goehring2011polarization}, phase transitions~\cite{hoffmann2012ginzburg}, and chemical waves~\cite{kuramoto2003chemical}.

In this example, we consider the $\lambda-\omega$ reaction-diffusion system as a prototypical model of  chemical waves~\cite{kopell1973plane}, showing how it can be inferred from data when including symmetry \emph{priors}. The model equations for the concentration fields $u(x,y,t)$ and $v(x,y,t)$ of two chemical species in 2D are:
\begin{align}\label{eq:RD}
    \frac{\partial u}{\partial t} &= D_u \left(\frac{\partial^2 u}{\partial x^2} + \frac{\partial^2 u}{\partial y^2}\right) + \lambda(r) u - \omega (r) v ,\\
    \frac{\partial v}{\partial t} &= D_v \left(\frac{\partial^2 v}{\partial x^2} + \frac{\partial^2 v}{\partial y^2}\right) + \omega(r) u  - \lambda(r) v.
\end{align}
Here, $r = \sqrt{u^2 + v^2}$, $\omega = -\beta r^2$, and $\lambda = 1 - r^2$. This system is symmetric in the two species, i.e., swapping $u \leftrightarrow v$ leaves the model unchanged. Such symmetries are common in biology and can be found in predator-prey models~\cite{freedman1980deterministic}, models of fish scale patterns~\cite{yang2002spatial}, and models of antagonistic protein interactions~\cite{goehring2011polarization}.

If known beforehand, such symmetries can be used as \emph{priors}. We impose the symmetry prior by grouping each column of the dictionary of one species with the corresponding column for the other species, where ``corresponding'' means pertaining to the same operator upon the swap, i.e., $uv^2 \leftrightarrow u^2v, u^2 \leftrightarrow v^2, u_{xx} \leftrightarrow v_{xx}$, etc. 

\begin{figure}[!t]
\centering
\includegraphics[width=3.4in]{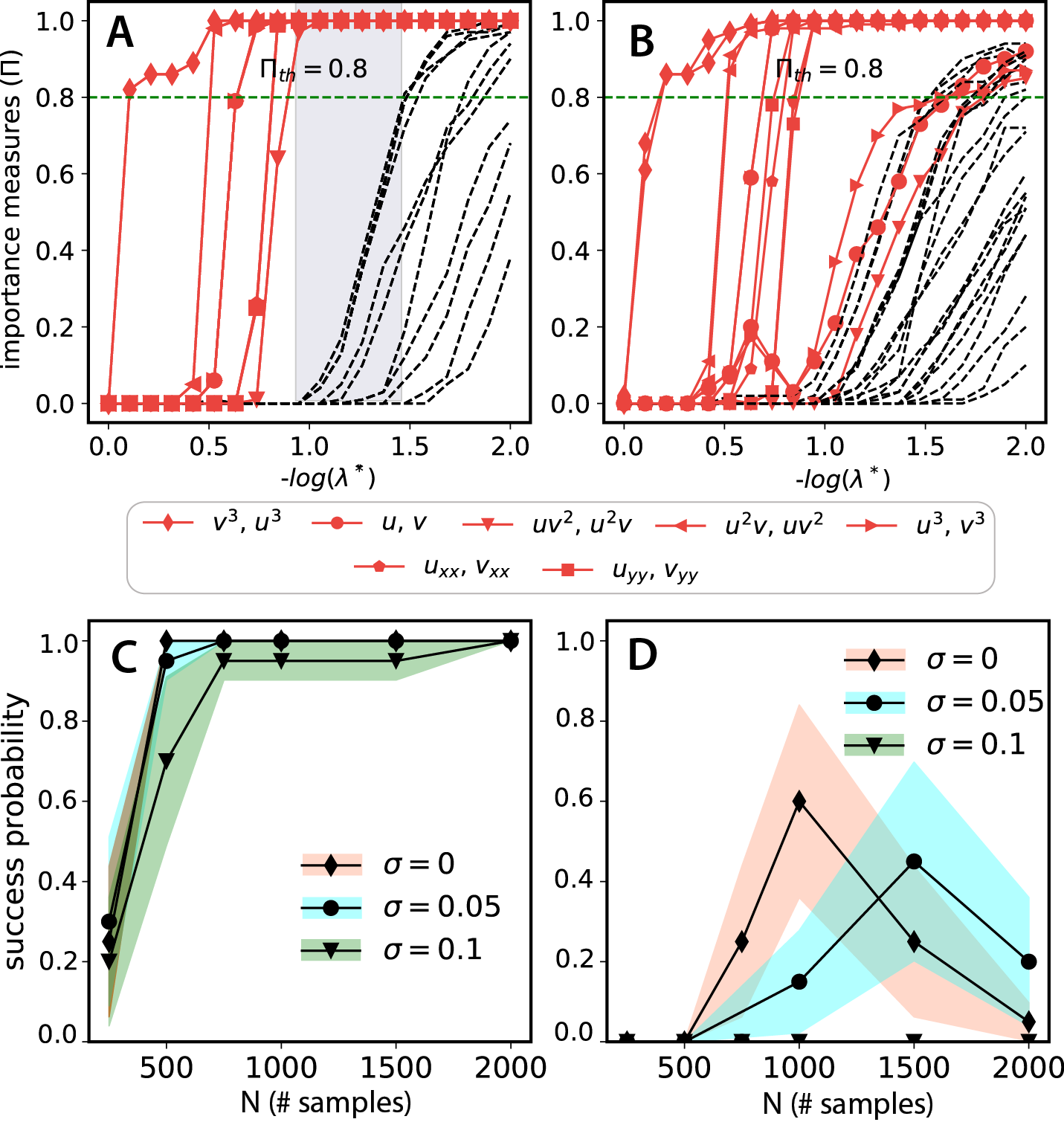}
\caption{\textbf{Inferring reaction-diffusion models from noisy spatio-temporal data.} \footnotesize{ 
\textbf{A, B}) Stability plots for with (A) and without (B) symmetry priors for noise level $\sigma=0.1$. \textbf{C,D} Achievability plots for with (C) and without (D) symmetry \emph{prior} for different noise levels in the data. Each point is averaged over 20 independent trials. The colored bands correspond to the Bernoulli variance. The block dictionary for each species is composed of nonlinearities upto 3rd order and spatial derivatives upto 2nd order ($p = 18$).}}
\label{fig:stability_plots}
\end{figure}

We use data obtained by numerically simulating the above model with 10\% point-wise Gaussian noise added to the data. The stability and achievability plots when using the data from Fig.~\ref{rd-data} are shown in Fig.~\ref{fig:stability_plots}. Comparing Fig.~\ref{fig:stability_plots}A,B, we observe that model inference without the symmetry \emph{prior} fails, whereas it works robustly when the prior is included via group sparsity. This fact is substantiated by the achievability plots in Figure~\ref{fig:stability_plots}(C,D) for model inference with and without the \emph{prior}, respectively, for different noise levels $\sigma$ in the data. Our group-sparse regression formulation provides remarkable consistency for model recovery over a wide range of $\lambda$ values even at high noise levels of 10\%.

\section{Conclusion and Discussion}\label{sec:concl}
\noindent We have introduced a flexible and robust inference framework to learn physically consistent differential-equation models from limited amounts of noisy data. 
We used the concept of group sparsity to provide a flexible way of including modeling priors to render inference more robust. We combined this with the concept of stability selection for principled selection of regularization parameters in cases where the true model is not known. To efficiently solve the resulting regression problem, we introduced the novel group Iterative Hard Thresholding (gIHT) algorithm. 

We have benchmarked and demonstrated the use of this algorithm in examples of common mathematical models in biological physics. The examples covered ordinary differential equations (ODEs) and partial differential equations (PDEs) in 1D and 2D. They demonstrated how different types of \emph{priors} can be imposed using the concept of group sparsity: conservation laws, model equivalence, spatially varying latent variables, and symmetries. The results have shown that including such priors enables correct model inference from data containing 10 or even 15\% noise. Without the priors, the correct model could not be recovered in any of the presented cases. The achievability plots furthermore confirmed that relatively little data (few hundred space-time points) is sufficient to reliably and reproducibly learn the correct model when group-sparsity \emph{priors} are included. Without the \emph{priors}, model inference was inconsistent in all cases.

Importantly, stability selection converts the problem of fine-tuning the regularization parameter $\lambda$ to the easier problem of thresholding the importance measure ($\hat{\Pi}$). We argue that this is easier to do, as it  relates to an upper bound on the number of false positives one is willing to tolerate~\cite{meinshausen2010stability}, providing interpretability. Adopting such results to the group-sparse case would be very useful for real-world applications in order to guarantee reliability of the underlying model.

The concepts introduced here are independent of how the elements of the dictionary are constructed. Exploring more advanced dictionary constructions, such as integral formulations~\cite{schaeffer2017sparse} or weak formulations~\cite{reinbold2020using}, in conjunction with group sparsity and stability selection likely provides a promising future research direction. 

In its current form, however, our framework has a number of limitations. First, we only considered \emph{non-overlapping} groups, restricting each column of the dictionary to be part of at most one group. This is a limiting assumption, as it is not uncommon in physics or biology to simultaneously use multiple overlapping \emph{priors}. The more advanced concept of structured sparsity~\cite{Bach2011a} could provide a way to include \emph{overlapping} priors in future work. Second, we only showed how to include \emph{priors} about the structure of a model. If additionally one wants to impose \emph{priors} about coefficient values (e.g., {\em values} of diffusion constants, reaction rates, etc.), the framework would need to be extended to constrained group-sparse regression \cite{boyd2004convex}. Third, although, we have demonstrated robust data-driven inference of the model structure, estimates for the coefficient values can considerable deviate from ground truth (see Appendix~\ref{sec:coeff_est}).

Especially at high noise levels, these estimation errors likely stem from inaccurate spatial derivative approximations, as the polynomial differentiation schemes used here amplify noise. These issues can possibly be addressed in the future by combining our framework with Physics Informed Neural Networks (PINNs) \cite{raissi2019physics} or Gaussian processes \cite{raissi2017machine} for more robustly estimating the coefficients of the recovered model once the model structure is fixed. Such hybrid methods, combining the reconstruction abilities of physics-constrained neural networks with the robustness and consistency of sparse inference methods, may be particularly powerful for recovering spatio-temporal latent variables, such pressure or stresses, that cannot be directly measured in experiments.

\begin{acknowledgments}
\noindent This work was supported by the Deutsche Forschungsgemeinschaft (DFG, German Research Foundation) under Germany’s Excellence Strategy -- EXC-2068-390729961 -- Cluster of Excellence ``Physics of Life'' of TU Dresden, and by the Center for Scalable Data Analytics and Artificial Intelligence (ScaDS.AI) Dresden/Leipzig, funded by the Federal Ministry of Science and Education (BMBF).
\end{acknowledgments}

\vfill\eject
\vfill\eject
\appendix
\counterwithin{figure}{section}

\section{Regression estimates of the coefficients}\label{sec:coeff_est}

\begin{figure}[h!]
\centering
\includegraphics[width=3.4in]{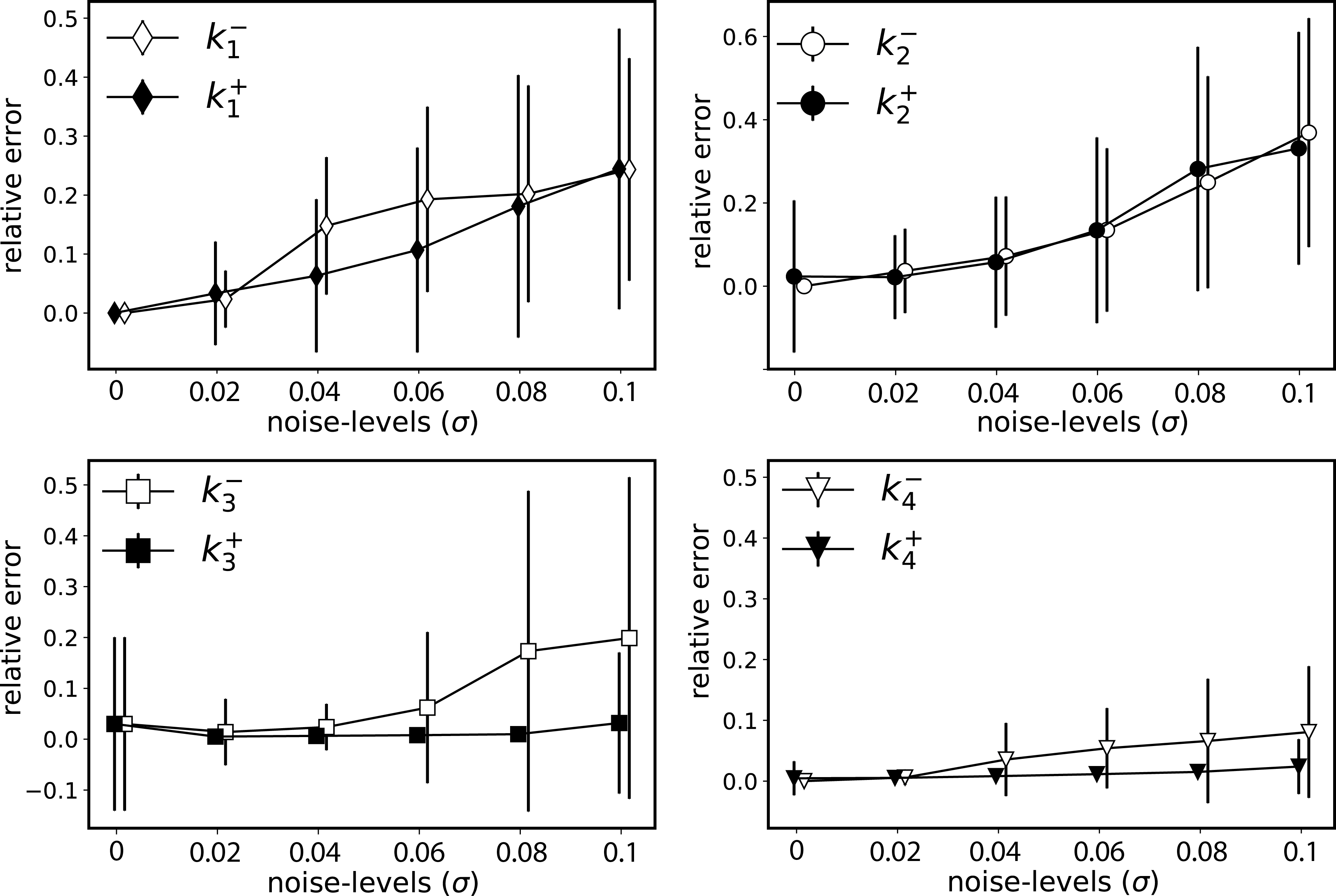}
\caption{\textbf{Relative errors in the coefficients inferred for JAK-STAT pathway reactions.} \footnotesize{ The plots show the estimates of the reaction rate constants of the JAK-STAT pathway as inferred by the gIHT algorithm for varying noise levels $\sigma$. The dashed line corresponds to the relative error $\frac{\vert \xi - \xi^* \vert }{\vert\xi^*\vert}$, where $\xi^*$ is the ground-truth value. The ground-truth values used here are: $k^\pm_1 = 0.021$, $k^\pm_2 = 2.46$, $k^\pm_3 = 0.2066$, and $k^\pm_4 = 0.10658$. 
The filled and unfilled symbols correspond to the estimated rate constants of different signs, which should be identical. 
}}
\label{fig:JK-STATcoeff}
\end{figure}

\begin{figure}[h!]
\centering
\includegraphics[width=3.4in]{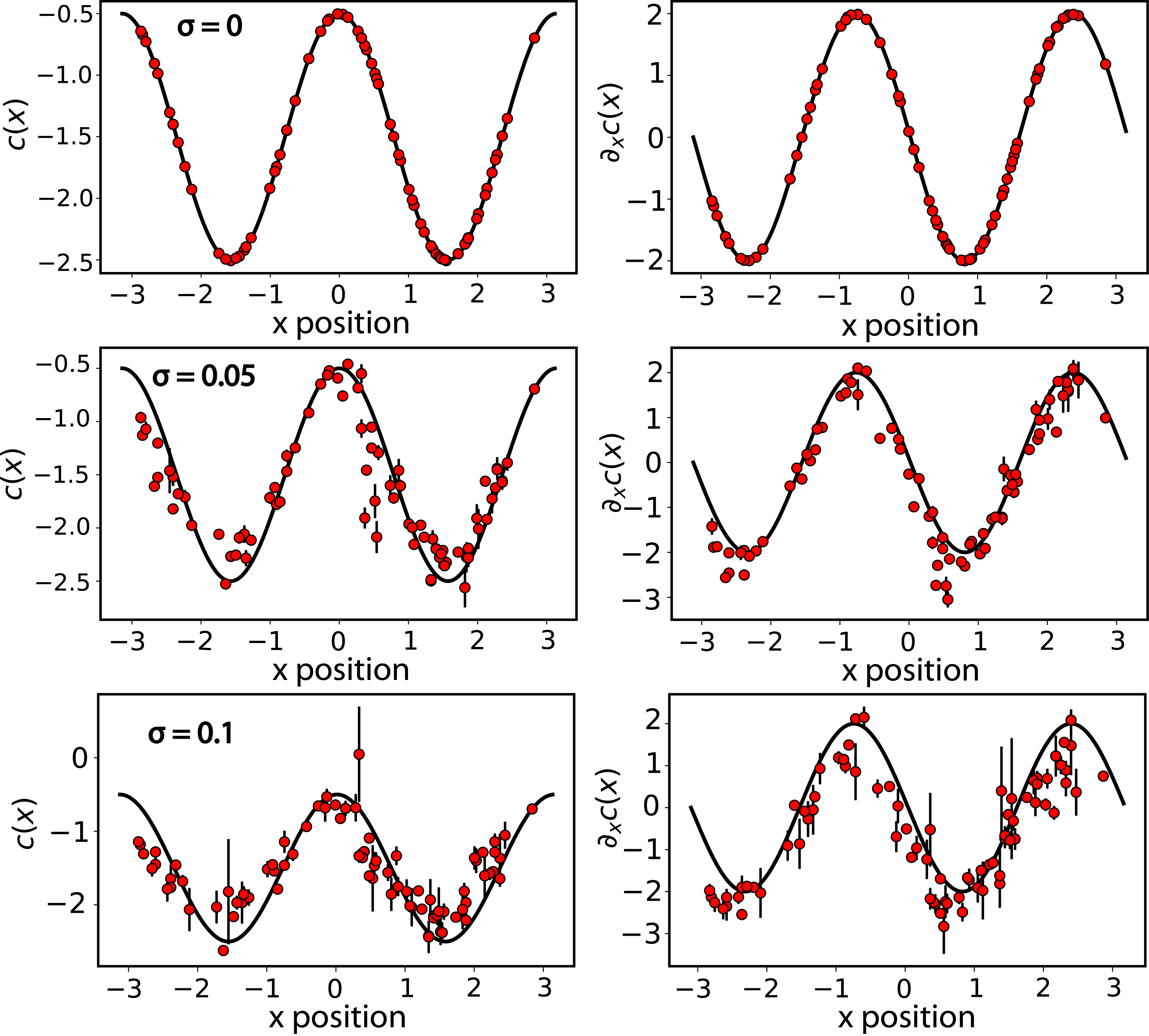}
\caption{\textbf{Spatially varying velocity field and its gradient for the advection-diffusion example.} \footnotesize{ The plots show the estimates for the latent spatially varying velocity $c(x)$ (left column) and its gradient $\partial_x c(x)$ (right column) from the gIHT algorithm. The rows correspond to the inference from data with different noise levels $\sigma$ (shown also in the inset). Symbols show estimated means with bars indicating estimation standard deviation over 20 independent trials. Solid black lines are ground truth.}}
\label{fig:PARprofiles}
\end{figure}

\begin{figure}[h!]
\centering
\includegraphics[width=3.4in]{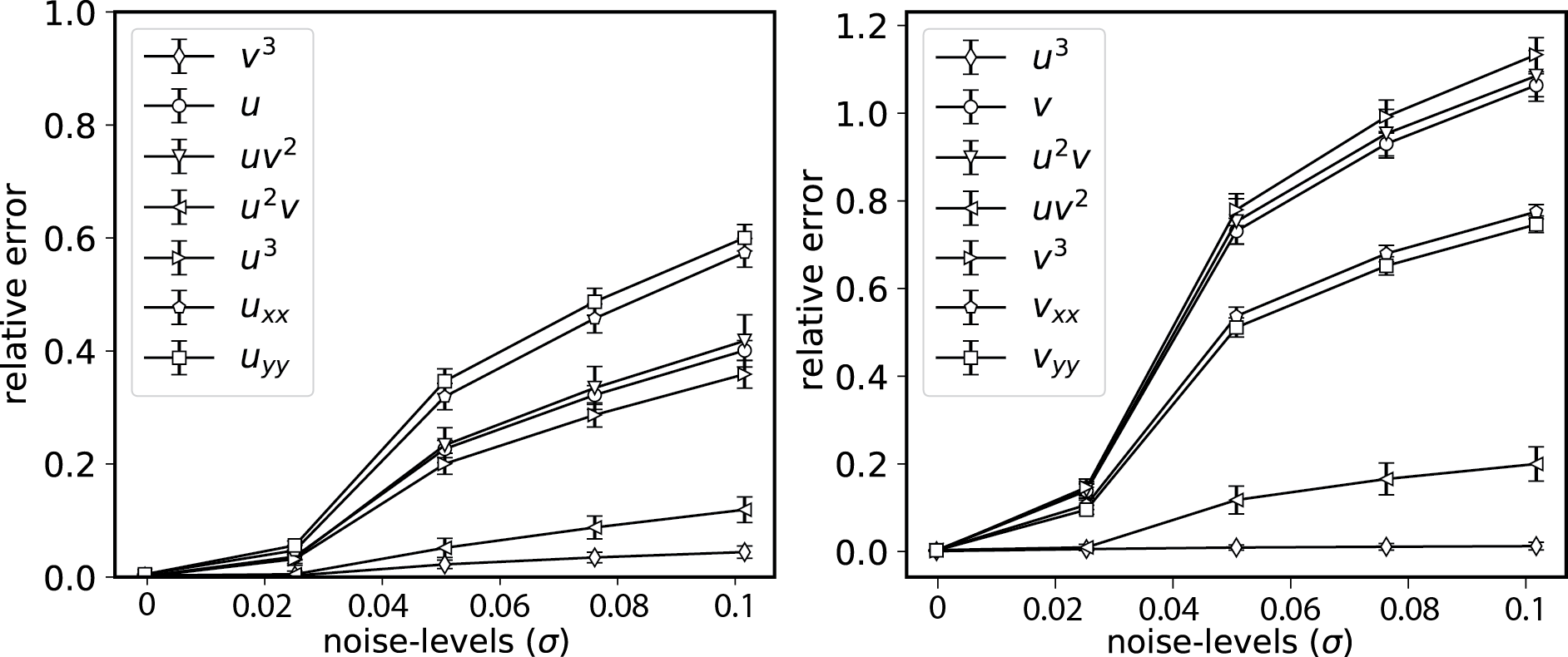}
\caption{\textbf{Relative errors in the coefficient estimation for the $\lambda-\omega$ reaction-diffusion system.} \footnotesize{ 
The plots shown the relative errors $\frac{\vert \xi - \xi^{*} \vert}{\xi^{*}}$ (vs.~ground truth) in the gIHT estimates of reaction coefficients and diffusion constants for the species $u$ (left) and $v$ (right) as a function of the noise level $\sigma$ in the data. The ground truth coefficients for the species $u$ and $v$ are as given in Eq.~(\ref{eq:RD}).}}
\label{fig:RDcoeff}
\end{figure}

\noindent The coefficients estimated by the gIHT algorithm from the noisy simulation data in the three application cases are shown in Figs.~\ref{fig:JK-STATcoeff} (for the JAK-STAT example), \ref{fig:PARprofiles} (for the advection velocity), and \ref{fig:RDcoeff} (for the reaction-diffusion system). In all cases, results are compared with ground-truth values for different noise levels. 

\section{Algorithm for group-sparse regression}

\noindent We derive and provide the details of the novel group Iterative Hard Thresholding (gIHT) algorithm presented here. The algorithm is based on an approximate proximal operator for \emph{non-overlapping} group sparsity, i.e., for cases where the groups $\{g_l : l \in \mathbb{N}_m\}$ form a partition of the index set $\mathbb{N}_p$.
In this case, the approximate proximal operator can be applied to each group separately, and the results summed~\cite{argyriou2011efficient}.

\subsection{Proximal view of the Iterative Hard Thresholding algorithm (IHT)}
\noindent We start from the well-known Iterative Hard Thresholding (IHT) algorithm for $\ell^0$-regularized sparse regression \cite{blumensath2009iterative}.
We formulate this algorithm from the perspective of projection and proximal operators. For solving the composite optimisation problem of Eq.~(\ref{general_opt}), we use  linearization and solve the following surrogate problem to generate a sequence $\{ \bm{\xi}^k \}$ as
\begin{align}\label{eq:iterates}
    \bm{\xi}^{k+1} = \arg\min_{\bm{\xi}} h(\bm{\xi}^k) & + \langle \nabla h (\bm{\xi}^k), \bm{\xi} - \bm{\xi}^k\rangle \\  &+  \frac{t^k}{2} \Vert \bm{\xi} - \bm{\xi}^k\Vert^2 + g (\bm{\xi}). \nonumber
\end{align}
This linearization works under the assumption that the loss function $f(\bm{\xi})$ is continuously differentiable with Lipschitz-continuous gradient, i.e., there exists a positive constant $\beta$ such that $\Vert \nabla f(\bm{x}) - \nabla f(\bm{y}) \Vert \leq \beta \Vert \bm{x} - \bm{y} \Vert \,\,\, \forall \bm{x},\bm{y} \in \mathbb{R}^d$. The problem in Eq.~(\ref{eq:iterates}) is equivalent to the proximal operator
\begin{equation}\label{eq:proxform}
    \textbf{prox}_g(\bm{v}^k) = \bm{\xi}^{k+1} = \arg\min_{\bm{\xi}} \Big\{ l(\bm{\xi}) \equiv \frac{1}{2} \Vert \bm{\xi} - \bm{v}^k \Vert^2 + g(\bm{\xi}) \Big\},
\end{equation}
where $\bm{v}^k = \bm{\xi}^k - \nabla h(\bm{\xi}^k)/t^k$  is the gradient-descent iterator. Thus, we perform gradient descent along $-\nabla h(\bm{\xi}^k)$ and then apply the proximal operator. In the Iterative Hard Thresholding (IHT) algorithm with non-convex penalty function $g(\bm{\xi}) = \lambda \Vert \bm{\xi}\Vert_0$, the proximal operator $\textbf{prox}_g(\bm{v})$ is approximated by hard thresholding~\cite{blumensath2009iterative}. 

\subsection{The \textit{approximate} proximal operator for the non-overlapping group sparsity problem}
\noindent We note that the above alternating gradient/proximal step is similar to the forward/backward splitting (FBS) algorithm~\cite{combettes2011proximal}. We therefore propose to use approximate thresholding also for the non-convex group sparsity problem.

The proximal operators for proper lower semi-continuous functions $g$ are well defined with the set $\textbf{prox}_{g}^{\lambda}$ being non-empty and compact~\cite{zhang2016alternating}. By extension of the idea of using thresholding as an approximation to the proximal step, we decompose the separable optimization problem in Eq.~(\ref{eq:proxform}) into a sum of sub-problems~\cite{argyriou2011efficient} and apply the approximate proximal operator (i.e., thresholding) to each sub-problem separately. For \emph{non-overlapping} groups, we can decompose the function $l(\bm{\xi})$ defined in Eq.~(\ref{eq:proxform}) into two parts:
\begin{align}\label{eq:separable}
l(\bm{\xi}) = & \left( \frac{1}{2} \Vert \bm{\xi}_{{g}_i} - \bm{v}_{{g}_i} \Vert_{2}^{2} + \lambda  \sqrt p_{{g}_i} I \left ( \Vert \bm{\xi}_{{g}_i} \Vert_{2} \neq 0 \right) \right)\\ \nonumber
 & + \left( \frac{1}{2} \Vert \bm{\xi}_{\bar{g}_i} - \bm{v}_{\bar{g}_i} \Vert_{2}^{2} + \lambda  \sum_{ j \neq  i}  \sqrt p_{\bar{g}_j} I \left ( \Vert \bm{\xi}_{\bar{g}_j} \Vert_{2} \neq 0 \right) \right),
\end{align}
where $\bar{g}_i = \{1,2,\ldots ,p\} - g_i$ is the complementary set of the group $g_i$. Since we restrict ourselves to non-overlapping groups, $ g_i \cap \bar{g}_i = \emptyset \,\,\,\forall i\neq j$. For a fixed $\bm{\xi}_{\bar{g}_i} = \bm{\xi}_{\bar{g}_i}^{*}$, it can be verified that $\Vert \bm{\xi}_{g_i}^{*} \Vert_2 = 0$ minimizes both terms in Eq.~(\ref{eq:separable}) if $\Vert \bm{v}_{g_i}\Vert \leq \sqrt{\lambda \sqrt{p}_{g_i}}$. For more details, we refer to \textit{Lemma 2} for the zero groups (i.e., for $\bm{\xi}_{g_i}^{*} = 0$) in the group LASSO problem~\cite{yuan2011efficient}. Similar arguments can be made for separable forms other than that shown in Eq.~\ref{eq:separable}, based on which we can formulate the  thresholding rule to minimize the function $l(\bm{\xi})$: 
\begin{equation}\label{eq:groupthresh}
H_{\textrm{group}}^{\lambda}(\bm{v}_g) =
\begin{cases} 
      0 & \text{if }  \Vert \bm{v}_g \Vert_2 < \sqrt{  \lambda \sqrt{p_g} } \\
      \bm{v}_g & \text{if }  \Vert \bm{v}_g \Vert_2 \geq \sqrt{  \lambda \sqrt{p_g} }.
   \end{cases}
\end{equation}

For group size $p_g=1$, this thresholding rule reduces to the popular Hard Thresholding (HT) algorithm, and the sequence $\{\bm{\xi}^k\}$ are iterates of the Iterative Hard Thresholding (IHT) algorithm~\cite{blumensath2009iterative, maddu2019stability}. Based on the generalized thresholding rule in Eq.~(\ref{eq:groupthresh}), we propose the following group Iterative Hard Thresholding (gIHT) algorithm with an additional de-biasing step~\cite{figueiredo2007gradient,foucart2011hard}.\\
\medskip
\begin{algorithm}[H]
\caption{group Iterative Hard Thresholding (gIHT) with de-biasing}
    \begin{algorithmic}[1]
    \renewcommand{\algorithmicrequire}{\textbf{Input:}}
 \renewcommand{\algorithmicensure}{\textbf{Output:}}
\REQUIRE $\bm{\Theta}, \bm{U}_t, \lambda, \mathcal{G}, \textrm{maxiter=10000}$
\vspace{0.5em} \ENSURE  $\hat{\bm{\xi}}$ \\
 \vspace{0.5em}
 \STATE Initialization : $\bm{\xi}^{0} = 0$
 \vspace{0.5em}
 \FOR {$k = 1$ to $\textrm{maxiter}$}
 \vspace{0.5em}
  \STATE $ \bm{v} = \bm{\xi}^{k} - \mu_t \nabla g (\bm{\xi}^{k})$
  \vspace{0.5em}
  \STATE $ \bm{u}^{k+1,1} = \textbf{H}_{\textrm{group}}^{\lambda}(\bm{v}), \:\:\: S^{k+1} = \textrm{supp}(\bm{u}^{k+1,1})$
  \vspace{0.5em}
  \STATE De-biasing: $\bm{\xi}^{k+1} = \arg \min_{\xi} \left(  \Vert \bm{U}_t - \bm{\Theta} \bm{\xi}  \Vert_{2}^{2} \right)_{S^{k+1}}$ 
  \vspace{0.5em}
  \ENDFOR
   \end{algorithmic}
\end{algorithm}



\bibliography{apssamp}

\end{document}